\pdfoutput=1

\documentclass[11pt]{article}

\usepackage[final]{acl}

\usepackage{times}
\usepackage{latexsym}

\usepackage[T1]{fontenc}

\usepackage[utf8]{inputenc}

\usepackage{microtype}

\usepackage{inconsolata}

\usepackage{graphicx}

\usepackage{xspace}
\usepackage{adjustbox}
\usepackage{booktabs}
\usepackage{multirow}
\usepackage{amsmath}
\usepackage{pifont}

\newcommand{\eg}{\emph{e.g}\xspace}
\newcommand{\ie}{\emph{i.e}\xspace}

\newcommand{\etc}{\emph{etc}\xspace}
\newcommand{\task}[0]{{\fontfamily{txtt}\selectfont {{VCD}}}\xspace}
\newcommand{\dataset}[0]{{\fontfamily{txtt}\selectfont {{{VCD}}}}\xspace}
\newcommand{\model}[0]{{\fontfamily{txtt}\selectfont {{VCM}}}\xspace}

\newcommand\imagenetvc{\textsc{ImageNetVc}\xspace}
\newcommand{\inlineimg}[2][1em]{\raisebox{-0.2\height}{\includegraphics[width=#1]{#2}}}

\title{\dataset{}: A Dataset for Visual Commonsense Discovery in Images}

\author{Xiangqing Shen, Fanfan Wang, Siwei Wu, and Rui Xia\thanks{Corresponding author.} \\
        School of Computer Science and Engineering, \\ Nanjing University of Science and Technology, China \\
        \texttt{\{xiangqing.shen, ffwang, wusiwei, rxia\}@njust.edu.cn}}

\begin{document}

\maketitle

\begin{abstract}

Visual commonsense plays a vital role in understanding and reasoning about the visual world. 
While commonsense knowledge bases like ConceptNet provide structured collections of general facts, they lack visually grounded representations. 
Scene graph datasets like Visual Genome, though rich in object-level descriptions, primarily focus on directly observable information and lack systematic categorization of commonsense knowledge. 
We present Visual Commonsense Dataset (\dataset{}), a large-scale dataset containing over 100,000 images and 14 million object-commonsense pairs that bridges this gap. 
\dataset{} introduces a novel three-level taxonomy for visual commonsense, integrating both Seen (directly observable) and Unseen (inferrable) commonsense across Property, Action, and Space aspects. 
Each commonsense is represented as a triple where the head entity is grounded to object bounding boxes in images, enabling scene-dependent and object-specific visual commonsense representation.
To demonstrate \dataset{}'s utility, we develop \model{}, a generative model that combines a vision-language model with instruction tuning to discover diverse visual commonsense from images.
Extensive evaluations demonstrate both the high quality of \dataset{} and its value as a resource for advancing visually grounded commonsense understanding and reasoning. 
Our dataset and code will be released on \url{https://github.com/NUSTM/VCD}.
\end{abstract}

\section{Introduction}
\label{sec:intro}



Commonsense, comprising facts and principles humans rely on in daily life, is essential for decision-making and behavior. Integrating it into AI systems enhances human-like reasoning, improves interpretability, and has become a growing area of research.
Visual commonsense, a portion of commonsense, refers to general knowledge about the visual world.
While existing commonsense knowledge bases, \eg, ConceptNet~\citep{conceptnet}, include some visual commonsense represented in textual form, they lack visually grounded commonsense—specific, contextually rich knowledge tied to actual visual scenes. 
 Such limitation results in restricted coverage and insufficient detail for effectively bridging vision and language understanding.
Cognitive science research~\citep{the_reviewing_of_object_files} indicates that humans perceive the world by focusing on objects in a scene, noting their attributes, spatial relationships, and actions to gather multidimensional information.
This information forms the basis of visual commonsense, which is inherently scene-dependent and object-specific.

\begin{figure*}[t]
  \centering
   \includegraphics[width=1.\linewidth]{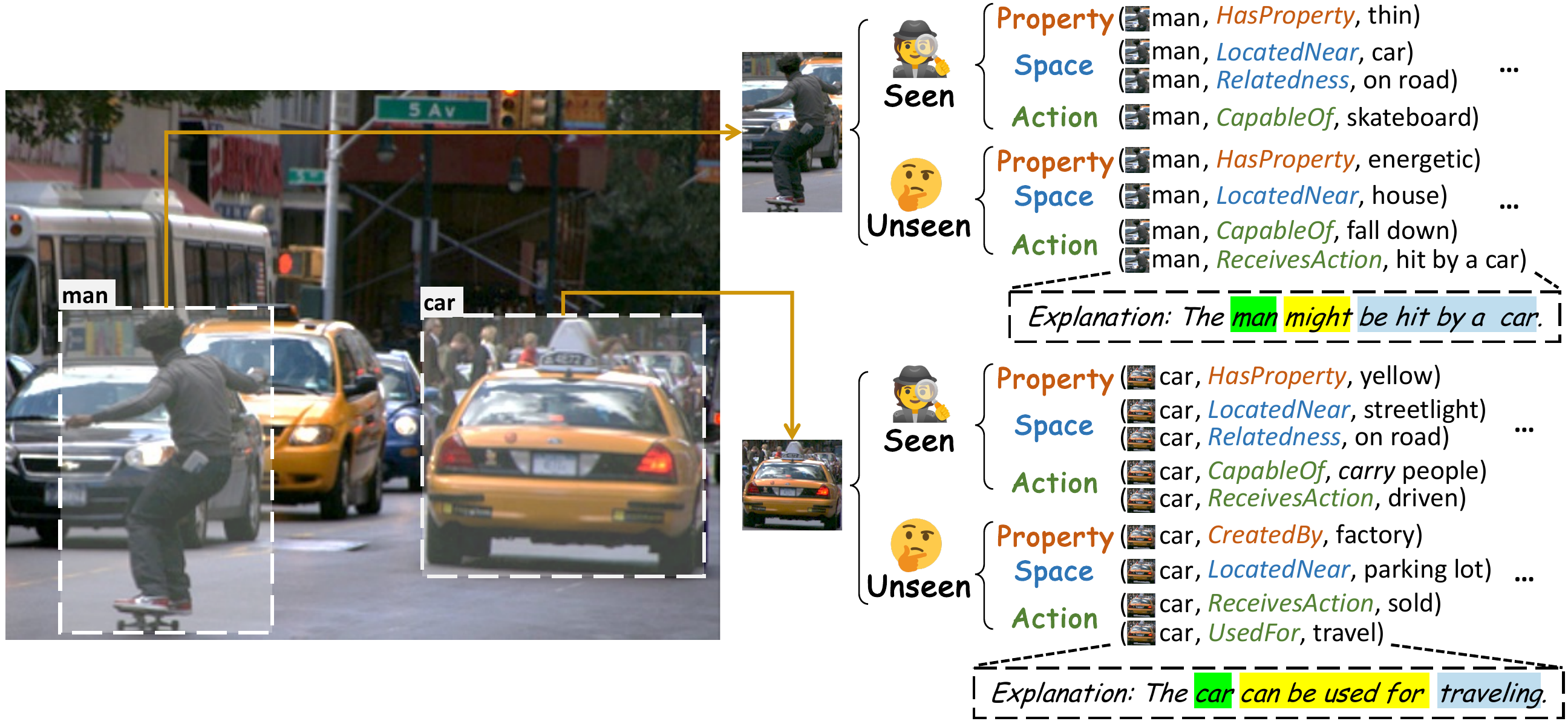}
   \caption{An example from \dataset{}. Given the left image, two objects (a man and a car) are annotated along with their associated 11 visual commonsense triples. These triples are organized within a hierarchical taxonomy. For example, (\inlineimg{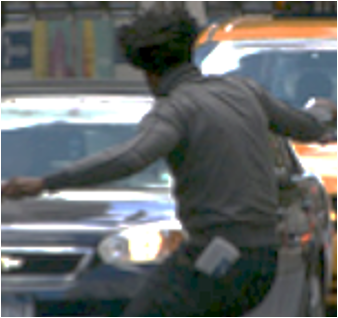}man, \textit{LocatedNear}, car) is a Seen commonsense under the Space aspect, 
(\inlineimg{figures/man_in_text.png}man, \textit{ReceivesAction}, hit by a car) is an Unseen commonsense
under the Action aspect.}
   \label{fig:motivation}
\end{figure*}

On the other hand, scene graph datasets in the field of computer vision, \eg, Visual Genome (VG) ~\citep{visual_genome}, although provide rich object-level descriptions of attributes, actions, and relationships, typically lack a systematic categorization of commonsense. 
Moreover, they predominantly focus on commonsense directly observable in images (referred to as \textit{seen commonsense} in this paper), while neglecting commonsense not visually apparent but still relevant to the image and can be inferred by general world knowledge (referred to as \textit{unseen commonsense}).
For example, in Fig.~\ref{fig:motivation}, given a scene depicting ``\textit{a man skateboarding on a busy street}'', humans can naturally infer unseen commonsense like ``\textit{the man might be hit by a car}''.
Such unseen commonsense is crucial for deep visual understanding and reasoning, but has received insufficient attention in current research.

To address these challenges, we present \dataset{}, a large-scale \underline{\textbf{V}}isual \underline{\textbf{C}}ommonsense \underline{\textbf{D}}ataset by integrating and linking Visual Genome and ConceptNet.
\dataset{} includes over 100,000 images with more than 14 million object-commonsense pairs, where each image is annotated with objects it contains, and each object is further annotated with its related visual commonsense triples.
Similar to ConceptNet, each commonsense is represented as a (head, relation, tail) triple; but the head here is a language-vision pair, consisting of an entity and its corresponding bounding box in the current image. 
By grounding commonsense entities to the bounding boxes in the image, \dataset{} bridges the gap between linguistic knowledge and visual information. 

We introduce a three-layer taxonomy to categorizing these visual commonsense triples.
First, we identify visually relevant categories from the 34 basic knowledge types defined in ConceptNet as the foundational layer.
These categories are then grouped into three fundamental aspects widely studied in computer vision (\ie, Property, Action, and Space), constituting the second layer.
For the top layer, we distinguish commonsense knowledge based on its visual observability in the given image (\ie, Seen commonsense versus Unseen commonsense).
This hierarchical taxonomy provides a comprehensive framework for organizing visual commonsense knowledge, bridging NLP and CV domains while enabling analysis of both observable and inferential visual relationships.

\dataset{} captures rich patterns and relationships that reflect the visual world, enabling the discovery of scene-dependent and object-specific visual commonsense.
In this regard, we train a generative model, \model{}, that integrates a vision-language model with instruction tuning, to generate visual commonsense from images.
The instructions cover diverse types of commonsense within the taxonomy, enabling \model{} to generate different categories of commonsense triples according to the provided instruction, spanning both Seen and Unseen visual commonsense across the Property, Action, and Space aspects.


Extensive evaluations, including both automatic and human evaluations, demonstrate 1) the high quality of the VCD dataset, 2) the strong performance in visual commonsense discovery, particularly surpassing GPT-4o in identifying unseen commonsense, and 3) the enhancement of downstream vision-language tasks through the discovered visual commonsense knowledge. 
These comprehensive evaluations demonstrate \dataset{}'s value as a foundational resource for discovering and leveraging visual commonsense, advancing visually-grounded commonsense AI.


\section{Related Work}
\label{sec:related_work}

Commonsense in text has been a longstanding research focus, with early studies primarily dedicated to constructing commonsense knowledge bases.
ConceptNet~\cite{conceptnet} integrates multiple knowledge bases.
ASER~\cite{DBLP:conf/www/ZhangLPSL20} captures selectional preference knowledge extracted from over 11 billion tokens of unstructured text.
TransOMCS~\cite{DBLP:conf/ijcai/ZhangKSR20} employs linguistic graphs to align ASER with ConceptNet.
DISCOS~\cite{DBLP:conf/www/FangZWSH21} enhances commonsense of ASER by aggregating information from neighboring concepts.
ATOMIC~\cite{sap2019atomic, DBLP:conf/aaai/HwangBBDSBC21, dense_atomic} is a collection of if-then knowledge triplets centered on daily events.

Research on visual commonsense has evolved from focusing on the specific commonsense category to broader, more generalized commonsense categories.
Early work focused on different specific dimensions of images, including taxonomy~\citep{neil}, unary affordance~\citep{mining_semantic}, physical properties~\citep{piglet, intrinsic_physical_concepts}, and spatial relationships~\citep{stating_the_obvious, automatic, acquiring_common_sense_spatial, grounding_consistency}.
More recent studies have expanded beyond these individual dimensions to explore generalized visual commonsense knowledge~\citep{learning_commonsense_through_visual_abstraction, hybird, things_not_written_in_text, visual_commonsense_reasoning, visual_commonsense_in_pretrained, can-language-models, viphy, imagenetvc}.
A crucial aspect of visual commonsense is the generation of scene graphs, which models object interactions within an image to support high-level reasoning~\cite{visual_genome, visual_relation_detection}.
Another important research direction involves multimodal knowledge graphs~\citep{onoro2017answering, ferrada2017imgpedia, liu2019mmkg, alberts2020visualsem, wang2020richpedia},
which extend traditional knowledge graphs by associating entities with non-textual data, such as images.
However, no existing multimodal knowledge graphs are explicitly designed to capture visual commonsense.
We distinguish visual recognition from visual commonsense, aligning the latter's ``seen'' aspects with ViCor's~\citep{zhou-etal-2024-vicor} Visual Commonsense Understanding (VCU). While visual recognition identifies objects and attributes (\textit{e.g.}, a ``man,'' ``thin''), VCU, or ``seen'' commonsense, provides an explicit, structured understanding of this literal visual content, such as ``(man, HasProperty, thin)'' or ``Person washing dishes''. Therefore, visual commonsense leverages visual recognition to build a structured, queryable layer of knowledge about directly observable elements and their explicit relationships within a scene, forming a foundational step towards deeper reasoning.

\begin{figure*}[t]
  \centering
   \includegraphics[width=1\textwidth]{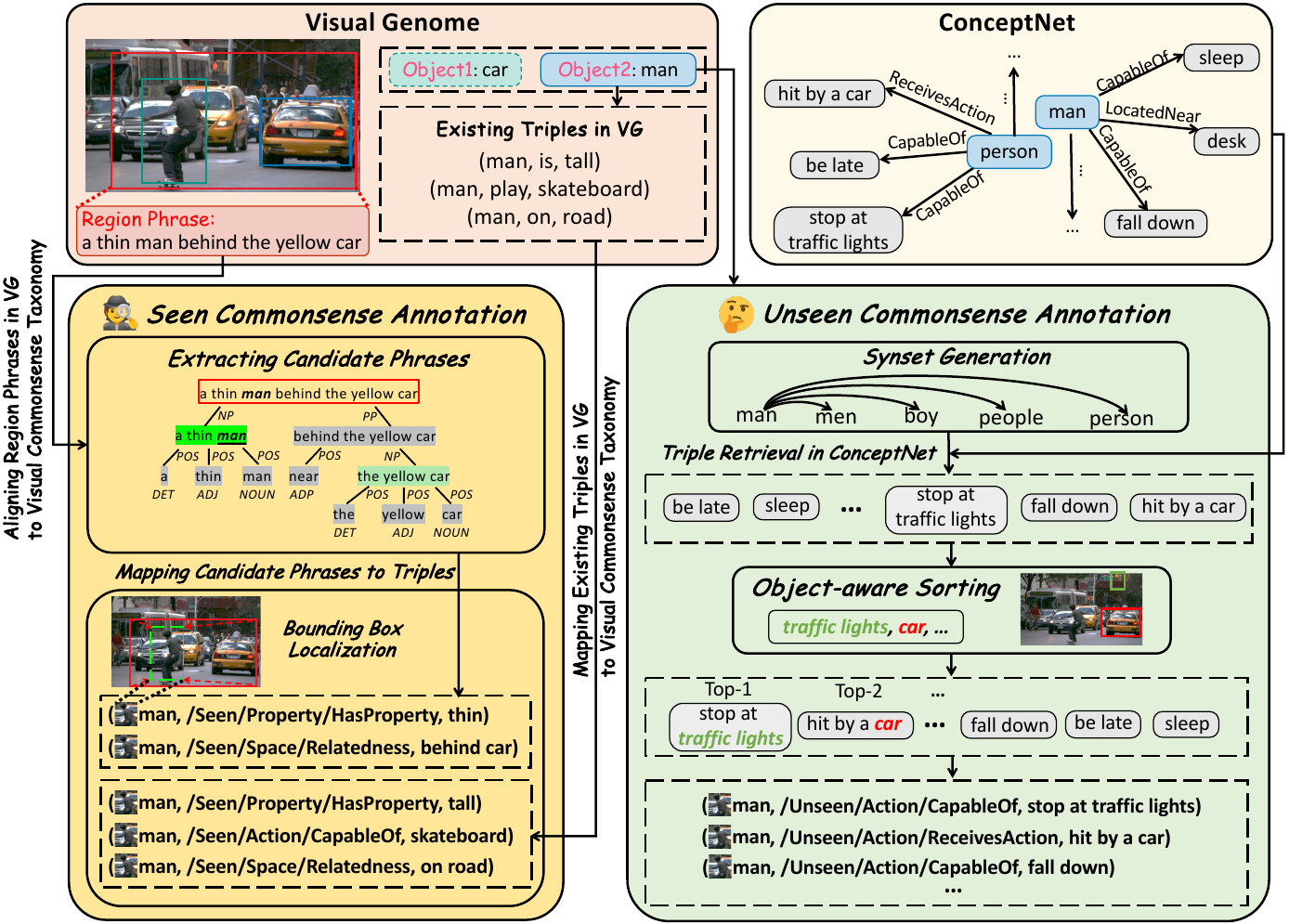}
   \caption{The construction process of \dataset{}.}
   \label{fig:dataset}
\end{figure*}


\section{Visual Commonsense Dataset Construction}

\label{sec:dataset_construction}

\subsection{Preliminary Resources}

ConceptNet~\citep{conceptnet} is a multilingual commonsense knowledge base that comprises a vast collection of manually curated triples, each representing words or phrases and their commonsense relationships.
It systematically defines 34 categories of commonsense and encompasses more than 4 million English triples.
However, its textual representation limits its ability to effectively capture scene-dependent and object-specific visual commonsense, which is crucial for understanding real-world contexts.

Visual Genome (VG)~\citep{visual_genome} is a large-scale scene graph dataset containing 108,077 images with dense annotations of 5.4 million region descriptions, 3.8 million object instances, 2.8 million attributes, and 2.3 million relations. 
Despite its extensive coverage, VG lacks a structured categorization of visual commonsense and does not include unseen commonsense.

\subsection{Visual Commonsense Taxonomy}
We introduce a three-layer hierarchical taxonomy
to organize the visual commonsense.
At first, we identify visually relevant categories from 34 basic knowledge types defined in ConceptNet, establishing the foundational layer.
These categories are then grouped into three fundamental aspects commonly employed in computer vision (\ie, Property~\citep{intrinsic_physical_concepts}, Action~\citep{mining_semantic}, and Space~\citep{acquiring_common_sense_spatial}), forming the second layer.
At the top layer, we introduce a visibility dimension that classifies knowledge as seen or unseen commonsense. 
Seen commonsense contains directly observable commonsense from images, 
While unseen commonsense involves inferred commonsense requiring contextual reasoning or life experiences.
This results in a hierarchical visual commonsense taxonomy.
Taking the image in Fig.~\ref{fig:motivation} as an example,
(\inlineimg{figures/man_in_text.png}man, \textit{LocatedNear}, car) is a Seen commonsense under the Space aspect, 
(\inlineimg{figures/man_in_text.png}man, \textit{ReceivesAction}, hit by a car) is an Unseen commonsense under the Action aspect.
More details of the visual commonsense taxonomy can be found in App.~\ref{sec:system}.

\subsection{Seen Commonsense Annotation}

VG encompasses a diverse range of real-world scenes, enriched with detailed annotations including object-level triples and region-level phrases, each accompanied by a bounding box.
These annotations make VG a valuable resource for capturing a broad spectrum of seen commonsense about various entities in an image by processing its existing object-level triples and region-level phrases.


Therefore, our approach to annotate seen commonsense is to map existing triples in VG to our visual commonsense taxonomy, and align region phrases in VG with the visual commonsense taxonomy.
\subsubsection{Mapping Existing Triples in VG to Visual Commonsense Taxonomy}

Fig.~\ref{fig:dataset} shows that VG includes objects marked with bounding boxes and annotated with descriptive triples.
These triples encapsulate visible attributes, actions, and spatial relationships.
For instance, (man, \textit{is}, tall) represents an attribute, while (man, \textit{play}, skateboard) and (man, \textit{on}, road) represent actions and spatial relationships, respectively.

To map these triples to seen commonsense categories, we establish part-of-speech based mapping rules.
Verbs are mapped to \textit{/Seen/Action}, adjectives to \textit{/Seen/Property}, and prepositions to \textit{/Seen/Space}.
For example, (man, \textit{is}, tall), as shown in Fig.~\ref{fig:dataset}, is mapped to (\inlineimg{figures/man_in_text.png}man, \textit{/Seen/Property/HasProperty}, tall) with the recognition of \textit{tall} as an \textit{adjective}.

Furthermore, \textit{/Seen/Space/LocatedNear} captures co-occurrence relationships, implying that two entities often appear together within the same visual scenario without a specific spatial relation.
For instance, since ``man'' and ``car'' co-occur in Fig.~\ref{fig:dataset}, one could infer a seen commonsense triple (\inlineimg{figures/man_in_text.png}man, \textit{/Seen/Space/LocatedNear}, car).

\begin{table*}[t]
\center
\caption{Comparison with other visual commonsense datasets. \# Categories represents the number of visual commonsense categories included in each dataset.} 
\begin{adjustbox}{max width=1.\linewidth}
\begin{tabular}{@{}lcccccc@{}}
\toprule
                            & Seen & Unseen & \# Categories & \# Images & \# BBox  & \# Commonsense \\ \midrule
ConceptNet~\citep{conceptnet}             &    \ding{52}      &   \ding{52}       & 34            &    \ding{56}     &   \ding{56} & $\approx$4M           \\
Visual Genome~\citep{visual_genome}          &    \ding{52}      &      \ding{56}    & \ding{56}             &    $\approx$106K      & $\approx$4.1M  &       $\approx$5M         \\
SpatialCS~\citep{things_not_written_in_text}   &          \ding{52}        &   \ding{56}          & 1             & \ding{56}      &  \ding{56}  & 1224           \\
ViComTe~\citep{visual_commonsense_in_pretrained}                &          \ding{52}        &   \ding{56}          & 1             & \ding{56}       & \ding{56} & 11114          \\
VEC~\citep{can-language-models} &          \ding{52}        &   \ding{56}          & 2             & \ding{56}     &  \ding{56}   & 4090           \\
VIPHY~\citep{viphy}                  &   \ding{52}        &   \ding{56}        & 2            & \ding{56}       &  \ding{56} & $\approx$30K            \\
ImageNetVC~\citep{imagenetvc}             &          \ding{52}        &   \ding{56}          & 2             & \ding{56}     &  \ding{56}  & 4976           \\
\midrule
\dataset{}                    &          \ding{52}        &   \ding{52}          & 11            &      $\approx$106K     &  $\approx$2.4M   &  $\approx$14M       \\ \bottomrule
\end{tabular}
\end{adjustbox}
\label{table:comparison}
\end{table*}

\subsubsection{Aligning Region Phrases in VG to Visual Commonsense Taxonomy}

\label{ssec:mapping_region_phrases}

Only mapping existing triples in VG may lead to omissions.
However, region phrases in VG can supplement these triples.
For example, consider a region phrase ``\textit{a thin man behind the yellow car}'' from VG in Fig.~\ref{fig:dataset}.
This phrase implicitly contains several seen commonsense triples that are missing from the existing triples, such as (\inlineimg{figures/man_in_text.png}man, \textit{/Seen/Space/Relatedness}, behind car), (\inlineimg{figures/man_in_text.png}man, \textit{/Seen/Property/HasProperty}, thin).
Consequently, we extract additional triples from region phrases using an automatic process.
A manual review is then conducted to ensure the reliability and accuracy of the extracted triples.


\paragraph{Extracting Candidate Phrases}
We begin by applying constituency parsing to region phrases to extract candidate phrases, including preposition, verb, and noun phrases.
For example, given the region phrase ``\textit{a thin man behind the yellow car}'' in Fig.~\ref{fig:dataset},
constituency parsing classifies the entire phrase as a prepositional phrase, as well as ``\textit{a thin man}'' and ``\textit{the yellow car}'' as noun phrases.

\paragraph{Mapping Candidate Phrases to Triples}
Upon candidate phrases, dependency parsing is used to determine their syntactic structure.
Then, for each type of candidate phrase, we use a carefully-defined set of mapping rules to map syntactic structures to commonsense triples.
To illustrate, for the noun phrase ``\textit{a thin man}'' where ``man'' is the \textit{root}, we apply the mapping rule ``\textit{adjective + noun} \(\rightarrow\) (noun, \textit{/Seen/Property/HasProperty}, adjective)'' to yield the triple (\inlineimg{figures/man_in_text.png}man, \textit{/Seen/Property/HasProperty}, thin).
Full set of mapping rules is in App.~\ref{sec:rules}.

\paragraph{Bounding Box Localization}
To determine the bounding boxes for objects in triples extracted from regional phrases, we first compute the overlap ratio between the given region and the annotated bounding boxes.
We then filter out boxes with an overlap ratio below a predefined threshold, retaining only those that meet or exceed this criterion to form a candidate set.
Next, we match object names from the triples to the candidate set, preserving only those triples that have a unique correspondence, while discarding those with multiple matches or no valid match.
The resulting triples are then used to enhance seen commonsense triples of objects.

\subsection{Unseen Commonsense Annotation}

Unseen commonsense is essential for visual commonsense reasoning beyond direct visual perception. While Visual Genome (VG) focuses on seen commonsense directly observable in images, it lacks annotations for unseen commonsense that is not visually present. ConceptNet complements this by providing a rich source of unseen commonsense.

Therefore, our approach to annotating unseen commonsense is to extract relevant knowledge triples from ConceptNet that correspond to objects in the image, serving as unseen commonsense knowledge for the given image.
The process consists of the following steps:


\paragraph{Synset Generation}
To enhance the coverage of unseen commonsense retrieved, we first lemmatize the names of objects in VG for synset generation, as shown in Fig.~\ref{fig:dataset}.

\paragraph{Triple Retrieval in ConceptNet}
Using the generated synset, we retrieve ConceptNet for unseen commonsense for each object, ensuring a comprehensive collection of unseen commonsense associated with each identified object in an image.

\paragraph{Object-aware Sorting}
Humans naturally consider all objects within a scene when making associations.
As shown in Fig.~\ref{fig:dataset}, an image of a man skateboarding alongside many cars may evoke unseen commonsense that the man is at risk of being hit by a car.
This connection between ``man'' and ``car'' arises from their co-occurrence in the image.
Building on this cognitive process, we prioritize unseen commonsense that involves objects present in the image, as they are more intuitively derived from the visual context.
This object-aware sorting strategy ensures that retrieved commonsense aligns more closely with human reasoning process.

\begin{figure}[t]
  \centering
   \includegraphics[width=0.98\linewidth]{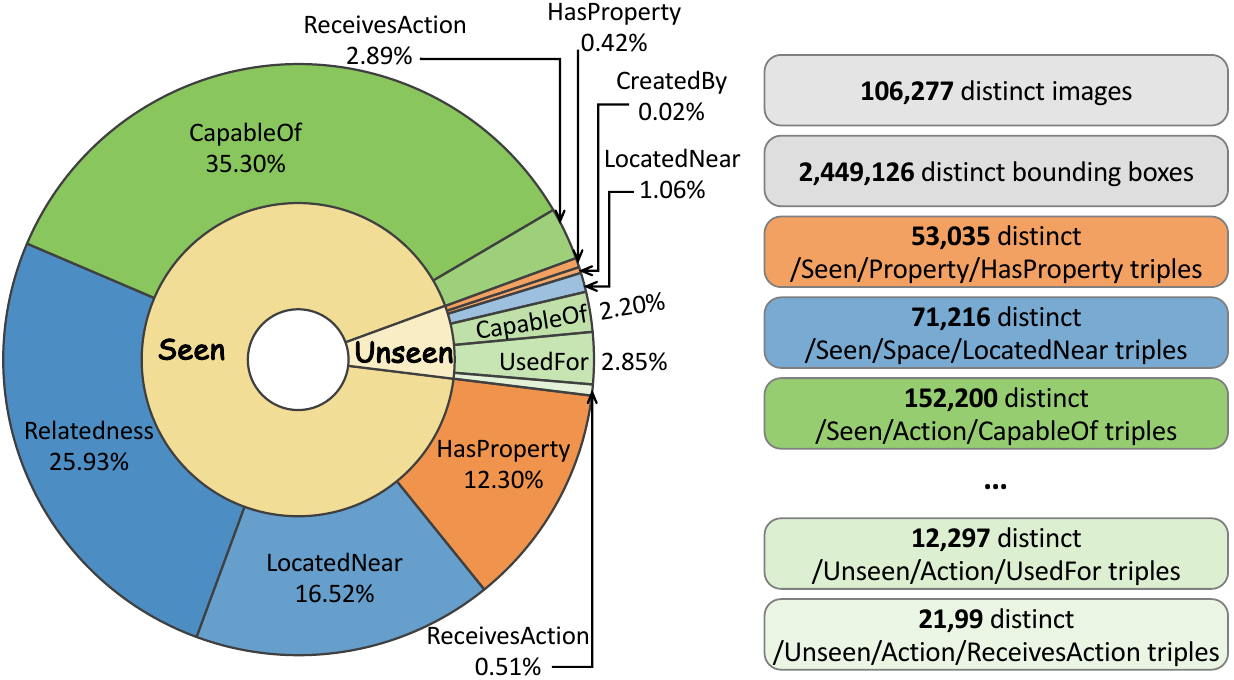}
   \caption{The statistics of \dataset{}.}
   \label{fig:stat}
\end{figure}
\subsection{Dataset Statistics}
Tab.~\ref{table:comparison} provides a comprehensive comparison of our dataset (\dataset{}) with existing visual commonsense datasets.
Unlike most datasets, \dataset{} integrates both seen and unseen commonsense, exhibiting distinct advantages in coverage, diversity, and scale.

Fig.~\ref{fig:stat} illustrates that \dataset{} consists of 106,277 images and 2,449,126 bounding boxes, encompassing 18,136 unique object names.
Furthermore, Fig.~\ref{fig:stat} presents the distribution of distinct commonsense triples across various categories, along with their respective proportions within \dataset{}.
Examples of \dataset{} are provided in App.~\ref{sec:examples}.

\begin{figure}[t]
  \centering
   \includegraphics[width=1.\linewidth]{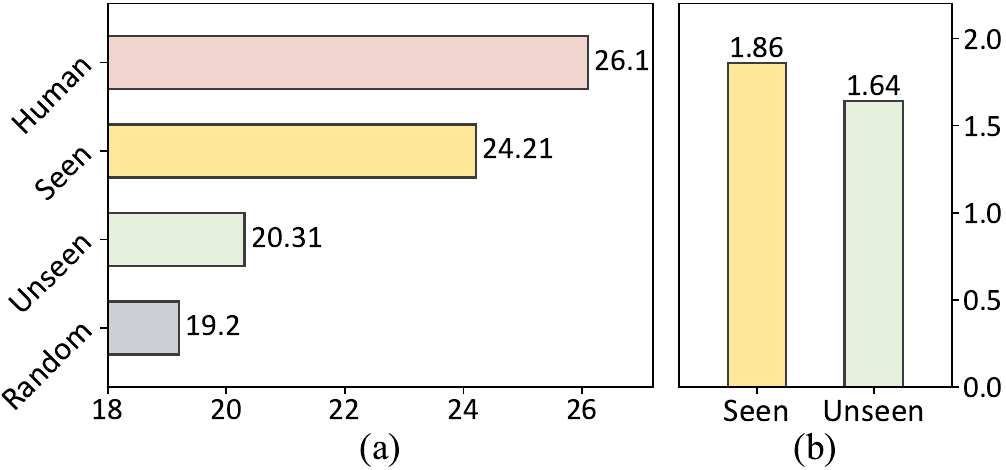}
\caption{(a) Automatic evaluation using CLIP similarity scores; (b) Human evaluation with Likert scale ratings.}
   
   \label{fig:data_clip_human}
\end{figure}

\subsection{Dataset Quality Control}

\dataset{} is built upon ConceptNet and VG, both of which are high-quality, manually annotated resources. While these foundations provide reliable base data, our work focuses on linking and aligning these resources. Therefore, we first evaluate the performance of our linking and alignment processes, followed by both automatic and human evaluations of the final annotations.

\paragraph{Evaluation of the Off-the-shelf Annotation Tools}
We utilize spaCy~\citep{spacy} for part-of-speech tagging and dependency parsing on region phrases in VG.
A human evaluation of 200 samples 
confirms a 99\% accuracy rate.
Similarly, for constituency parsing, AllenNLP~\citep{allennlp} achieves 97.5\% accuracy based on human evaluation of 200 samples.\footnote{The high accuracy of both spaCy and AllenNLP is largely due to simple linguistic structure of region phrases in VG.}

\paragraph{Evaluation of the Iterative Annotation Process}
To establish the rule set in Sec.~\ref{ssec:mapping_region_phrases}, we follow an iterative annotation process.
In each iteration, 200 samples are examined.
If the error rate exceeds 5\%, the rules are refined and reassessed.
This process repeats until the error rate falls below 5\%, ensuring that \dataset{} adheres to rigorous quality standards.

\paragraph{Automatic Evaluation of the Annotated Commonsense}
Following ~\citet{DBLP:conf/nips/GadreIFHSNMWGZO23,DBLP:conf/nips/SchuhmannBVGWCC22}, we evaluate the quality of \dataset{} by calculating image-commonsense matching scores using the CLIP model.
As shown in Fig.~\ref{fig:data_clip_human}(a), we derive a lower bound by randomly sampling commonsense triples for an image and computing the match score.
For the upper bound, we manually annotate ground truth seen commonsense for 300 images and calculate their matching scores.
Fig.~\ref{fig:data_clip_human}(a) shows that the seen commonsense matching score is slightly below the upper bound, indicating the high quality of \dataset{}.
The unseen commonsense matching score is only slightly higher than the lower bound, as these scores pertain to objects within the image but do not match the image semantically.

\paragraph{Human Evaluation of the Annotated Commonsense}
In addition to automatic evaluation, we perform human evaluation on 4,000 randomly selected images using a 0-2 Likert scale (higher is better),
with ratings provided by undergraduate students working on vision-language learning.
Following \citep{DBLP:conf/nips/Ouyang0JAWMZASR22}, we first assess their agreement with researcher-labeled examples and select the 10 evaluators with the highest agreement scores.
Fleiss's Kappa of 0.804 indicates a good agreement.
As shown in Fig.~\ref{fig:data_clip_human}(b), seen commonsense receives high preference, aligning with CLIP scores.
While unseen commonsense has lower CLIP scores, evaluators still favor it, indicating that it effectively reflects commonsense not depicted in the image.

\section{Visual Commonsense Discovery and Its Evaluation}

\label{sec:methods}

\dataset{} provides a comprehensive foundation for discovering visual commonsense from images, capturing rich patterns and relationships that enable the application of such knowledge to enhance downstream VL tasks.
Leveraging this dataset, we train \model{}, a \underline{\textbf{V}}isual \underline{\textbf{C}}ommonsense Discovery \underline{\textbf{M}}odel that combines a generative VL architecture with instruction tuning.
This allows \model{} to generate visual commonsense from novel images.
The generated visual commonsense can be used to improve the performance of downstream VL applications.\footnote{As this paper primarily focuses on \dataset{} construction,
we present a concise version of this part here due to space limitations, with more detailed descriptions in the Appendix.}

\subsection{Training a Visual Commonsense Discovery Model}

Given an image \(I\), an object \(o_{i}\) with a bounding box, and a commonsense category \(r_{k}\), \model{} aims to generate a set of \(m\) commonsense triples \(T_{i}^{k}\):
\begin{equation}
T_{i}^{k} = \{t_{1}, t_{2}, \ldots, t_{m}\} = \text{\model{}}(I, o_{i}, r_{k}),
\end{equation}
where \(o_{i} \in \{o_{1}, \ldots, o_{j} \}\) represents the set of objects identified within \(I\) (each annotated with a bounding box), \(r_{k} \in \{r_{1}, \ldots, r_{l}\}\) represents the set of all types of visual commonsense, \(t_{m} = (o_{i}, r_{k}, c)\)
is a commonsense triple, and \(c\) is in the form of nouns, adjectives, or phrases.

Additionally, we iterate over each object \(o_{i} \in \{o_{1}, \ldots, o_{j}\}\) and each type of visual commonsense \(r_{k} \in \{r_{1}, \ldots, r_{l}\}\), in order to discover a comprehensive set of commonsense triples $\mathcal{T}$:
\begin{equation}
\mathcal{T} = \bigcup_{i=1}^{j} \bigcup_{k=1}^{l} T_{i}^{k} = \{T_{1}^{1}, .., T_{1}^{l}, T_{2}^{1}, .., T_{j}^{l}\}.
\end{equation}

\begin{figure}[t]
  \centering
   \includegraphics[width=1.\linewidth]{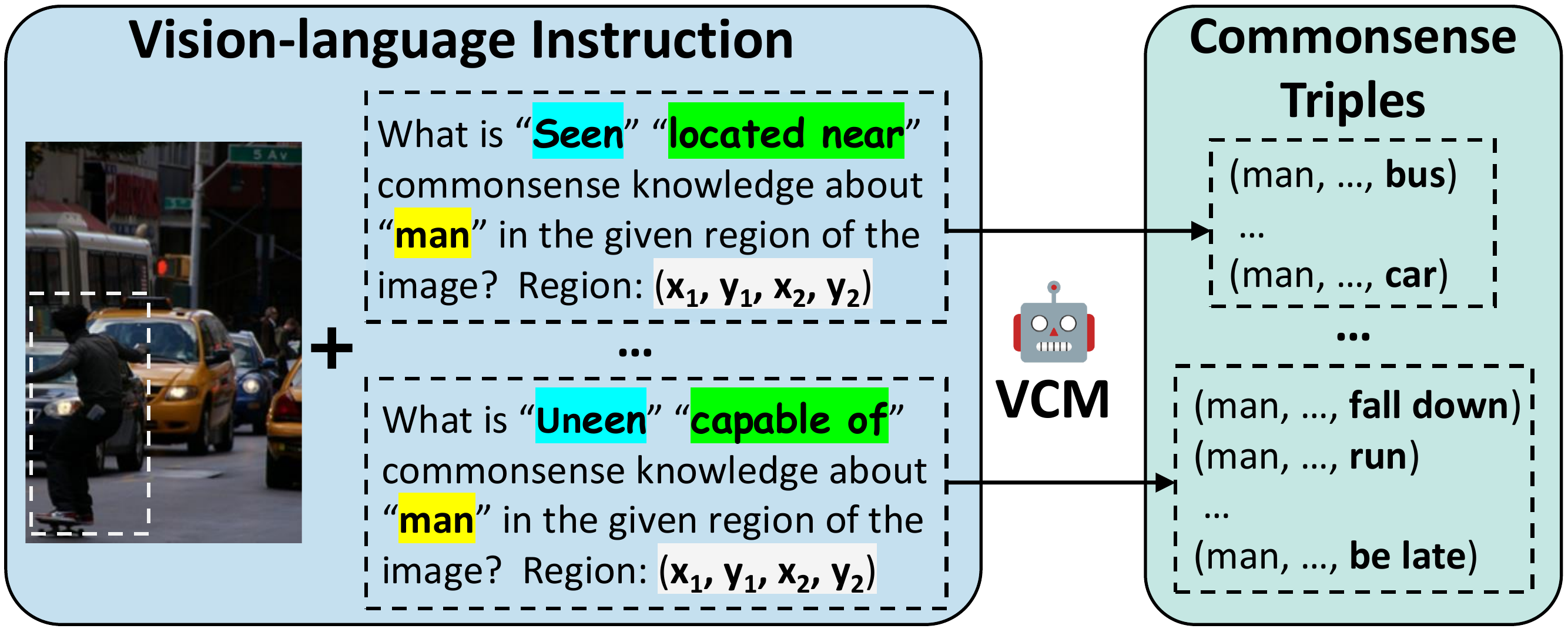}
   \caption{The input and output in \model{}.}
   \label{fig:model}
\end{figure}

The input of \model{} comprises an image, the name of an object with a bounding box, and a category of visual commonsense to discover.
As shown in Fig.~\ref{fig:model}, these versatile elements are integrated into one instruction template using instruction tuning methods~\citep{instruct_blip, multi_instruct}.
The output of \model{} is a series of commonsense triples generated in an autoregressive manner.
\model{} aims to minimize the following loss function:
\begin{equation}
\mathcal{L}=-\sum_{i=1}^{|y|} \log P_\theta\left(y_i \mid y_{<i}, x\right),
\end{equation}
where \(\theta\)  denotes model parameters, \(x\) represents the instruction and \(y\) denotes the commonsense.
The training details are provided in App.~\ref{sec:evaluation_of_downstream_tasks}

\begin{table}[t]
\center
\caption{Automatic evaluation results of \model{}.}
\begin{adjustbox}{max width=1\columnwidth}
\begin{tabular}{@{}lccccc@{}}
\toprule
Model & B-1$\uparrow$  & B-2$\uparrow$ & R$\uparrow$ & M$\uparrow$ & W$\downarrow$ \\
\midrule
SPHINX & 6.4 & 2.0 & 7.6 & 6.2 & 170.5 \\
Qwen-VL & 9.4 & 3.0 & 11.5 & 10.2 & 120.5 \\
GPT-4o  & 15.1 & 6.3 & 19.5 & 16.0 & 100.7 \\
OFA$_{\text{large}}$ & 1.1 & 0.3 & 2.0 &  1.2 &  100.8\\
\model{}$_{\text{tiny}}$  & 41.3 & 31.5 & 44.8 & 37.8 & 78.3 \\
\model{}$_{\text{medium}}$  & 48.8 & 37.6 & 52.4 & 45.2 & 73.5 \\
\model{}$_{\text{base}}$  & 53.9 & 42.4 & 56.8 & 50.1 & 71.0 \\
\midrule
\model{}$_{\text{large}}$  & 56.6 & 45.6 & 59.9 & 53.3 & 67.1 \\
\quad w/o image & 51.3 & 43.7 & 56.4 & 51.6 & 72.5 \\
\quad w/o region & 51.5 & 42.6 & 55.4 & 47.7 & 69.4 \\
\quad w/o name & 39.9 & 28.9 & 44.3 & 36.2 & 91.5 \\
\bottomrule
\end{tabular}
\end{adjustbox}
\label{table:main_results}
\end{table}

\subsection{Evaluation of Visual Commonsense Discovery}

\subsubsection{Evaluation Protocol}

\paragraph{Automatic Evaluation Metrics}
For the automatic evaluation of \model{}'s visual commonsense discovery capabilities, we employ metrics for natural language generation: BLEU-1 (B-1)~\citep{bleu}, BLEU-2 (B-2), Rouge-L (R)~\citep{rouge}, METEOR (M)~\citep{meteor}, and Word Error Rate (W)~\citep{wer}.

\paragraph{Human Evaluation Metrics}

While automatic evaluation metrics provide preliminary insights, they may not fully capture the diversity and quality perceived by humans.
Therefore, we conduct a human evaluation, focusing on the correctness and completeness of commonsense generated by \model{}.
For each type of visual commonsense, 10 images are randomly selected.
Their associated outputs are compared against those produced by multimodal large language models.
Two independent evaluators analyze the results, and any disagreements are resolved by a third.
The evaluations are structured as win/draw/lose comparisons.
Fleiss' kappa scores reveal moderate agreement among evaluators.
Evaluators receive salary at a rate of \$8 per hour, which is above the local average wage.

\subsubsection{Automatic Evaluation}

Tab.~\ref{table:main_results} summarizes the results of the automatic evaluation.
Upon analyzing the results in Tab.~\ref{table:main_results}, we can find that
\model{} exhibits improvements in all automatic evaluation metrics as the model scale increases, which is consistent with our expectations.
Particularly, \model{}$_{\text{tiny}}$ shows a significant decrease in performance, highlighting the importance of the model scale for \task{}.

\begin{table}[t]
\center
\caption{Comparison with MLLMs by human evaluation. W.R., D.R. and L.R. represent rate of win, draw and lose, respectively.}
\begin{adjustbox}{max width=1.\columnwidth}
\begin{tabular}{@{}lcccccc@{}}
\toprule
\multirow{2}{*}{~\model{}$_{\text{large}}$} & \multicolumn{3}{c}{Seen} & \multicolumn{3}{c}{Unseen} \\ \cmidrule(l){2-7} 
                      & W.R.      & D.R.    & L.R.    & W.R.      & D.R.    & L.R.    \\ \midrule
vs. OFA$_{\text{large}}$              & 100\%   & 0\%  & 0\%   & 100\%   & 0\%  & 0\%  \\                     
vs. \model{}$_{\text{base}}$              & 34\%   & 56\%  & 10\%   & 36\%   & 62\%  & 2\%  \\ \midrule
vs. SPHINX & 68\%   & 14\%  & 18\%  & 91\%     & 7\%    & 2\%    \\ 
vs. Qwen-VL & 59\%   & 29\%  & 12\%  & 86\%     & 11\%    & 3\%    \\ 
vs. GPT-4o              & 28\%   & 42\%  & 30\%  & 41\%     & 29\%    & 30\% \\ \bottomrule
\end{tabular}
\end{adjustbox}
\label{table:human_evaluation}
\end{table}

\subsubsection{Human Evaluation}

The results of the human evaluation are summarized in Tab.~\ref{table:human_evaluation}.
\model{}$_{\text{large}}$ consistently outperforms \model{}$_{\text{base}}$ across both seen and unseen commonsense discovery,
aligning with the findings of automatic evaluation and reinforcing the reliability of automated metrics.
Furthermore, \model{}$_{\text{large}}$ significantly outperforms SPHINX, which struggles to adhere to instructions, particularly in identifying and generating unseen commonsense.

While GPT-4o demonstrates strong performance in generating diverse examples of seen commonsense, it underperforms in unseen commonsense discovery, often confusing seen and unseen commonsense.
This suggests that GPT-4o faces challenges in distinguishing and generating them as distinct categories.
In contrast, \model{}$_{\text{large}}$ exhibits a stronger ability to associate unseen commonsense with image objects, leading to more precise unseen commonsense discovery.

\subsubsection{Ablation Study}

To evaluate the impact of images, bounding boxes, and object names on \model{}’s performance, we conduct ablation studies.
Results in Tab.~\ref{table:main_results} show that removing any of these elements degrades performance, with the removal of object names causing the most significant drop, even below \model{}$_\text{base}$.
This highlights the critical role of textual information from object names.

\subsection{Evaluation on Vision-language Tasks}
We further evaluate the effectiveness of visual commonsense discovered by \model{} on two downstream VL tasks.
The first is a dedicated evaluation of a model's visual commonsense capabilities, while the second involves visual question answering (VQA) datasets that require both image understanding and external knowledge.
Intuitively, the discovered visual commonsense can enhance performance on both VL tasks.
Due to space limitations, we provide more details in the Appendix.

\paragraph{Visual Commonsense Evaluation} We assess whether \dataset{} could enhance \model{}'s visual commonsense capabilities by comparing the backbone and \model{} on \imagenetvc{}, where we would expect \model{}, fine-tuned on \dataset{} upon the backbone, to demonstrate superior performance.
The experimental results are reported in Tab.~\ref{table:imagenetvc}.
It is observed that \model{}$_{\text{large}}$ shows improvements in the categories of Color, Material, Component, and Others, indicating an overall improvement in commonsense knowledge.
However, there is a minor decrease in recognizing Shapes, likely due to a deficiency in \dataset{}'s shape-related commonsense.

\begin{table}[t]
\center
\caption{Visual commonsense capacity on ImageNetVC.}
\begin{adjustbox}{max width=1.\columnwidth}
\begin{tabular}{@{}lcccccc@{}}
\toprule
    & \textsc{Col.} & \textsc{Sha.} & \textsc{Mat} & \textsc{Com.} & \textsc{Oth.} & \textsc{Avg}  \\ \midrule
OFA & 47.2  & 72.6  & 66.7   & 100.0   & 85.1   & 80.7 \\
\model{} & 56.6  & 69.3  & 73.5   & 99.7   & 88.1   & 83.5 \\ \bottomrule
\end{tabular}
\end{adjustbox}
\label{table:imagenetvc}
\end{table}

\paragraph{Visual Question Answering} This section evaluates on VQAv2~\citep{making} and OK-VQA~\citep{okvqa} to underscore the significance of the commonsense discovered by \model{} for VQA.
We compare the results from the backbone against those from the backbone incorporating commonsense discovered by \model{} as complement information for answering the question.
The experimental results are reported in Tab.~\ref{table:vqa}.
We can find that integrating the commonsense discovered by \model{} indeed enhances the performance of both VQA datasets, which is evidence of the significance of \dataset{} for downstream VL tasks.


\begin{table}[t]
\center
\caption{Significance of commonsense on VQA tasks.}
\begin{adjustbox}{max width=1.\columnwidth}
\begin{tabular}{@{}lcc@{}}
\toprule
     & VQAv2 & OK-VQA\\ \midrule
OFA &  75.3 & 33.8 \\
\hspace{1em} w/ commonsense   & 75.8  & 34.6 \\ \midrule 
Qwen-VL-7B & 79.5   & 58.6 \\
\hspace{1em} w/ commonsense & 79.9    & 60.0 \\ 
\bottomrule
\end{tabular}
\end{adjustbox}
\label{table:vqa}
\end{table}




\section{Conclusion}

We introduced \dataset{}, a large-scale Visual Commonsense Dataset that bridges the gap between linguistic and visual commonsense. By combining structured relations from ConceptNet with object-level annotations from Visual Genome, \dataset{} provides both seen and unseen commonsense across three aspects: Property, Action, and Space. This hierarchical taxonomy enables fine-grained, scene-dependent, and object-specific commonsense representation.
To demonstrate its utility, we developed \model{}, a generative model trained with instruction tuning. The model exhibits a strong ability in discovering various types of visual commonsense, and improves performance on vision-language tasks like visual question answering.
\dataset{} provides a foundation for enhancing visually grounded commonsense understanding and reasoning, enabling AI systems to better capture visual commonsense knowledge for real-world applications.

\section*{Limitations}


One limitation of our study is that, due to space constraints, this paper primarily focuses on introducing \dataset{} and the corresponding task of visual commonsense discovery.
Our evaluation of the significance of discovered visual commonsense for vision-language tasks is relatively simple.
In future work, we plan to conduct a more systematic and comprehensive evaluation across a broader range of downstream vision-language tasks.

Furthermore, this study considers only visual commonsense in static images, leaving out dynamic, temporal, and causal visual commonsense as reflected in videos.
Exploring these aspects presents a promising research direction which we aim to pursue in future work.

\section*{Acknowledgments}

This work was supported by the Natural Science Foundation of China (No. 62476134).



\bibliography{custom, mybib}

\newpage

\appendix

\section{Details of \dataset{} Construction}

\subsection{Definition and Examples for Visual Commonsense Taxonomy}
\label{sec:system}

First, we present the definition of categories of visual commonsense in Tab.~\ref{table:types}.
Then, based on Fig.~\ref{fig:motivation}, we furthermore provide examples for each category of visual commonsense as an illustration.
To maintain conciseness, not all bounding boxes in the image depicted in Fig.~\ref{fig:motivation} are annotated.

\begin{table*}[t]
\centering
\caption{Definitions and examples of visual commonsense taxonomy.}
\renewcommand{\arraystretch}{1.5} 
\begin{adjustbox}{max width=1.\textwidth}
\begin{tabular}{lllll}
\toprule
\multicolumn{3}{c}{\textbf{Category}} & \multicolumn{1}{c}{\textbf{Definition}} & \multicolumn{1}{c}{\textbf{Example}} \\\midrule
\multirow{5}{*}{Seen} & Property & HasProperty & The object has a \textit{currently seen} property, such as shape, color, or material. & (car, \textit{/Seen/Property/HasProperty}, yellow) \\
\cline{2-5} 
& \multirow{2}{*}{Space} & LocatedNear & The object co-occurs with another object in a \textit{currently seen} manner, without a specific spatial relationship. & (car, \textit{/Seen/Space/LocatedNear}, streetlight) \\
\cline{3-5} 
& & Relatedness & The object has a \textit{currently seen} spatial relationship with another object. & (car, \textit{/Seen/Space/Relatedness}, after a car) \\
\cline{2-5} 
& \multirow{2}{*}{Action} & CapableOf & The object performs a \textit{currently seen} active action. & (car, \textit{/Seen/Action/CapableOf}, drive on road) \\
\cline{3-5} 
& & ReceivesAction & The object undergoes a \textit{currently seen} passive action. & (skateboard, \textit{/Seen/Action/ReceivesAction}, played by man) \\\midrule
\multirow{6}{*}{Unseen} & \multirow{2}{*}{Property} & HasProperty & The object has a \textit{currently unseen} property, such as shape, color, or material. & (iron, \textit{/Unseen/Property/HasProperty}, hard) \\
\cline{3-5} 
& & CreatedBy & The object has a \textit{currently unseen} method of creation. & (car, \textit{/Unseen/Property/CreatedBy}, factory) \\
\cline{2-5} 
& Space & LocatedNear & The object co-occurs with another object in a \textit{currently unseen} manner, without a specific spatial relationship. & (man, \textit{/Unseen/Space/LocatedNear}, sofa) \\
\cline{2-5} 
& \multirow{3}{*}{Action} & CapableOf & The object performs a \textit{currently unseen} active action. & (man, \textit{/Unseen/Action/CapableOf}, grow up) \\
\cline{3-5} 
& & UsedFor & The object has a \textit{currently unseen} function or purpose. & (car, \textit{/Unseen/Action/UsedFor}, drive to work) \\
\cline{3-5} 
& & ReceivesAction & The object undergoes a \textit{currently unseen} passive action. & (car, \textit{/Unseen/Action/ReceivesAction}, hit) \\
\bottomrule
\end{tabular}
\end{adjustbox}
\label{table:types}
\end{table*}

\subsection{Full Set of Mapping Rules}
\label{sec:rules}

This section presents the full set of mapping rules, detailed in Tab.~\ref{table:mapping}.
Tab~\ref{table:mapping}'s first column displays the syntactic structures identified through syntactic parsing of noun phrases, which are the result of constituent syntactic analysis.
Depending on these syntactic structures, a subsequent mapping to a variety of types of visual commonsense is performed, as indicated in Tab.~\ref{table:mapping}'s third column, and is based on the parts of speech (POS) outlined in the second column.
To facilitate comprehension, Tab.~\ref{table:mapping} also includes corresponding examples and explanations for each distinct mapping rule.

\begin{table*}[ht]
\caption{Full set of mapping rules.}
\center
\begin{adjustbox}{max width=1.\textwidth}
\begin{tabular}{@{}cclll@{}}
\toprule
 &
  \textbf{POS} &
  \multicolumn{1}{c}{\textbf{Category}} &
  \multicolumn{1}{c}{\textbf{Example}} &
  \multicolumn{1}{c}{\textbf{Explanation}} \\ \midrule
PP &
  - &
  /Seen/Space/Relatedness &
  \textbf{man} \textit{before the yellow car} $\rightarrow$ (man, \textit{/Seen/Space/Relatedness}, before car) &
  \begin{tabular}[c]{@{}l@{}}``\textbf{Man}'' is the root noun. Regarding the prepositional phrase ``before the\\ yellow car'', we simplify it to ``before car'' to obtain the basic triple form.\end{tabular} \\ \midrule
\multirow{2}{*}{VP} &
  VBN &
  /Seen/Action/ReceivesAction &
  \textbf{man} \textit{hit by a yellow car} $\rightarrow$ (man, /Seen/Action/ReceivesAction, hit by a car) &
  \begin{tabular}[c]{@{}l@{}}``\textbf{Man}'' is the root noun. Regarding the verbal phrase ``hit by a yellow\\  car'', since POS of ``hit'' is ``VBN'', we map it to a passive action.\end{tabular} \\
 &
  VBG &
  /Seen/Action/CapableOf &
  \textbf{car} \textit{driving on the road} $\rightarrow$ (car, \textit{/Seen/Action/CapableOf}, driving on road) &
  \begin{tabular}[c]{@{}l@{}}``\textbf{Man}'' is the root noun. Regarding the verbal phrase ``driving on the\\  road'', since POS of ``driving'' is ``VBG'', we map it to a active action.\end{tabular} \\ \midrule
\multirow{2}{*}{NP} &
  ADJ &
  /Seen/Property/HasProperty &
  a small \textbf{car} $\rightarrow$ (car, \textit{/Seen/Property/HasProperty}, small) &
  \begin{tabular}[c]{@{}l@{}}``\textbf{Car}'' is the root noun. Regarding the noun phrase ``a small car'', since\\ POS of ``small'' is ``ADJ'', we map it to ``\textit{/Seen/Property/HasProperty}''.\end{tabular} \\
 &
  VBG &
  /Seen/Action/CapableOf &
  a running \textbf{man} $\rightarrow$ (man, \textit{/Seen/Action/CapableOf}, run) &
  \begin{tabular}[c]{@{}l@{}}``\textbf{Man}'' is the root noun. Regarding the noun phrase ``a running man'', since\\ POS of ``running'' is ``VBG'', we map it to ``\textit{/Seen/Action/CapableOf}''.\end{tabular} \\ \bottomrule
\end{tabular}
\end{adjustbox}
\label{table:mapping}
\end{table*}

\subsection{Examples in \dataset{}}
\label{sec:examples}

\begin{figure*}[t]
  \centering
   \includegraphics[width=1\linewidth]{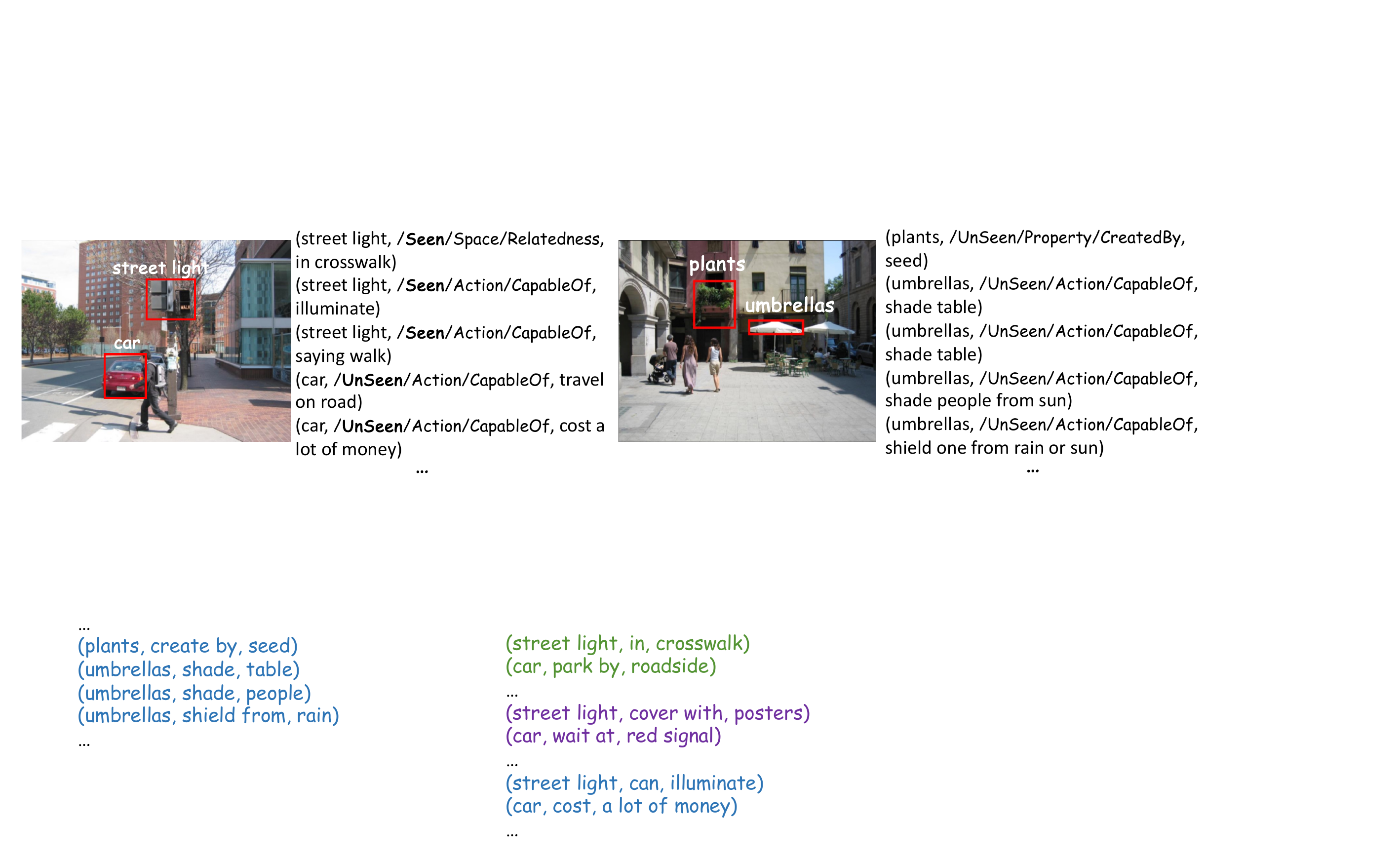}
   \caption{Examples in \dataset{}.}
   \label{fig:dataset_examples}
\end{figure*}

We provide examples from \dataset{} as illustrated in Fig.~\ref{fig:dataset_examples}.
\dataset{} includes detailed commonsense corresponding to each object within the image, delineated by bounding boxes.
This commonsense is expressed in the form of triples.
For the sake of conciseness, we do not provide the complete set of commonsense for every object annotated with bounding boxes in the image.

\subsection{Word Cloud for \dataset{}}

\begin{figure*}[t]
  \centering
   \includegraphics[width=1.\linewidth]{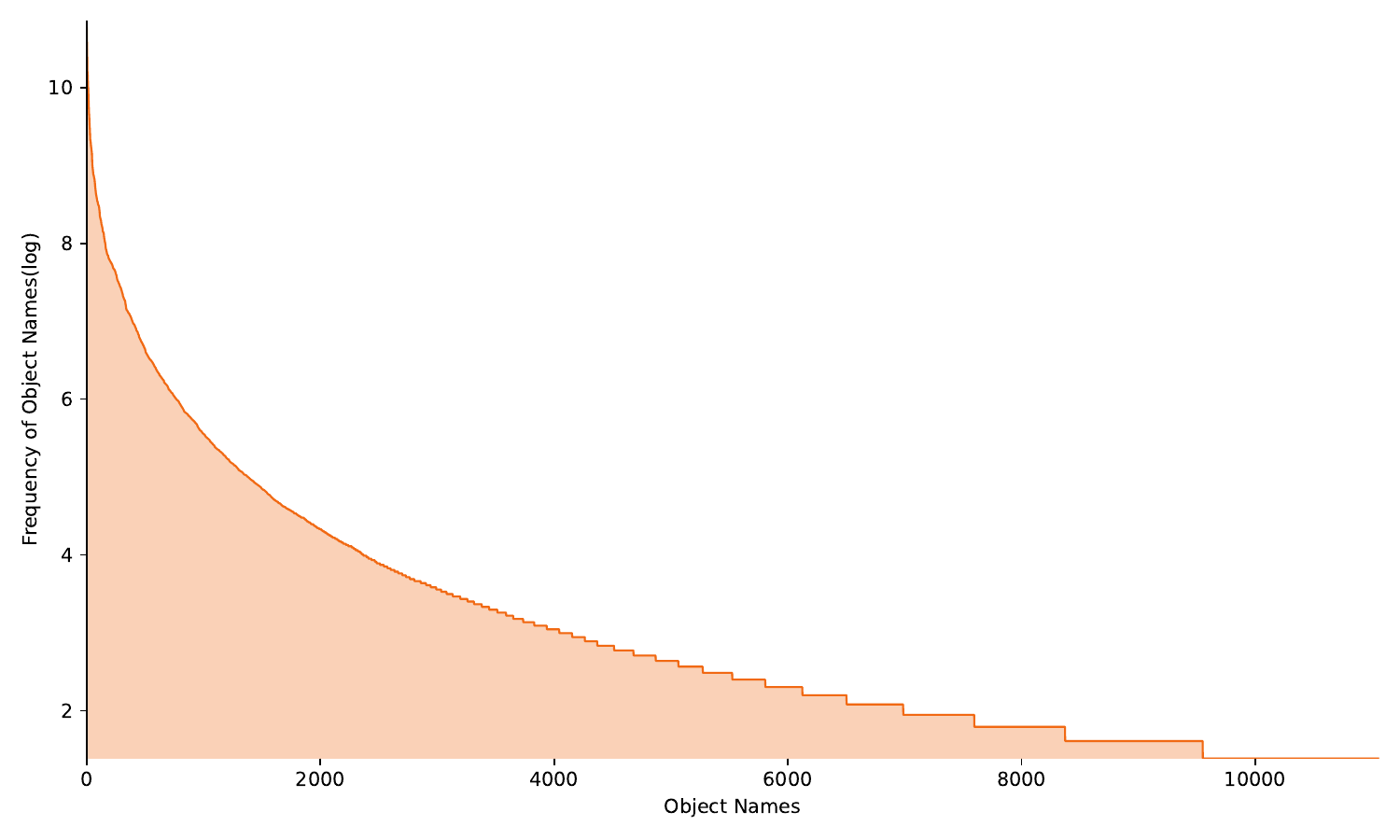}
   \caption{Distribution of object names in \dataset{}.}
   \label{fig:word_freq}
\end{figure*}

\begin{figure}[t]
  \centering
   \includegraphics[width=1\linewidth]{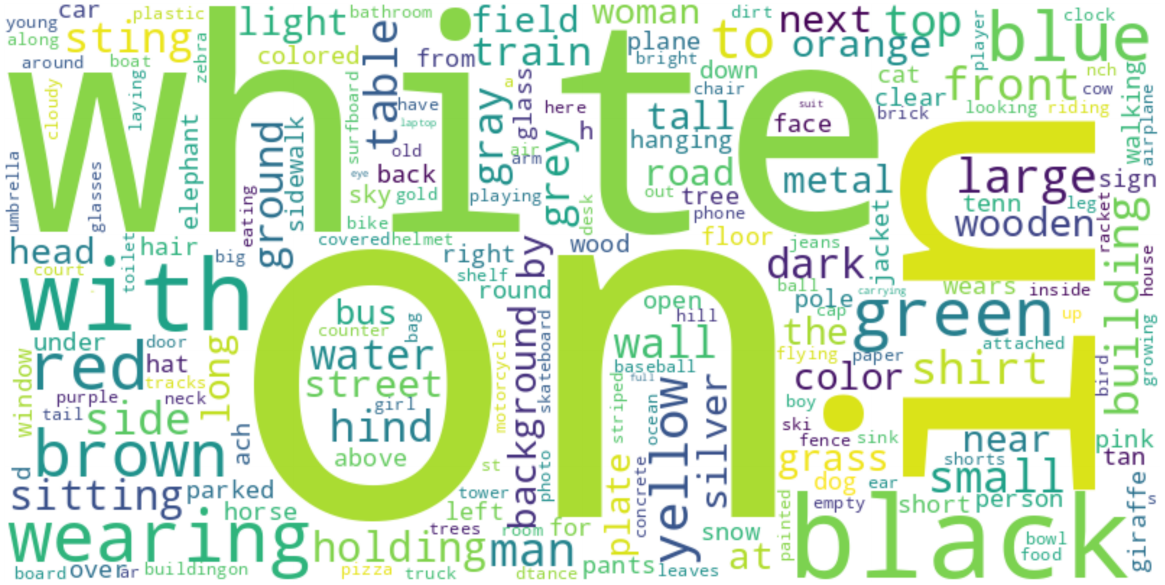}
   \caption{Word cloud for seen commonsense in \dataset{}.}
   \label{fig:word_ex}
\end{figure}

\begin{figure}[t]
  \centering
   \includegraphics[width=1\linewidth]{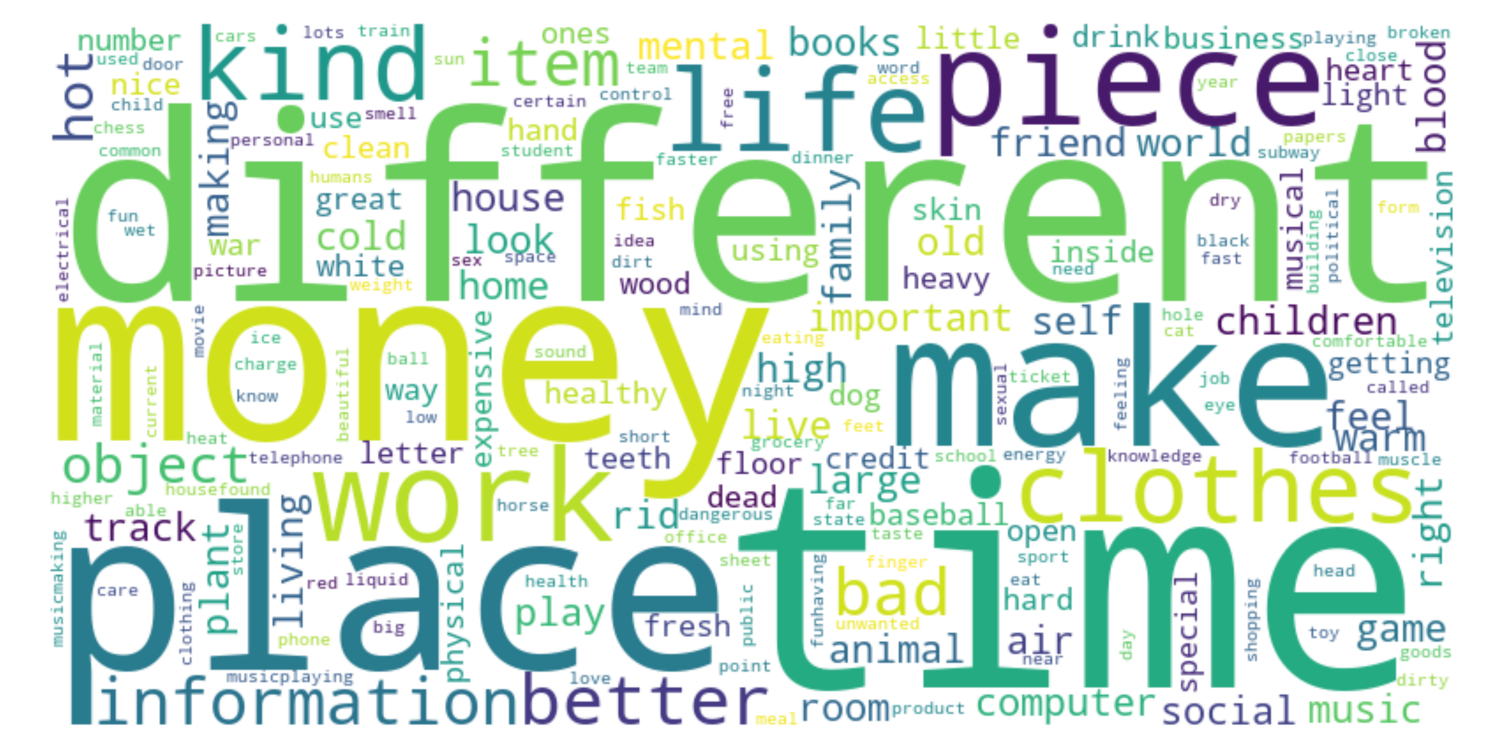}
   \caption{Word cloud for unseen commonsense in \dataset{}.}
   \label{fig:word_im}
\end{figure}

In Fig.~\ref{fig:word_ex} and Fig.~\ref{fig:word_im}, we provide word clouds for seen and unseen commonsense, repectively.
It can be observed that seen commonsense frequently contains words describing positions and colors, \eg, ``white'' and ``on''.
While in the word cloud for unseen commonsense, the word cloud contains words like ``time'' and ``place'', which are abstract concepts.
The differences well align to the definition of seen and unseen commonsense.
 
\section{Details of \model{} Training and Evaluation}
\label{sec:evaluation_of_downstream_tasks}

\subsection{Implementation Details of \model{}}
\model{} is based on OFA~\citep{ofa}, an encoder-decoder architecture.
While early vision-language models primarily focused on image-text alignment, more recent VL models have enhanced object localization capabilities by extending to region-text alignment.
However, few models effectively incorporate coordinate specifications within instructions.
We require a model with strong localization abilities that can seamlessly integrate coordinate information into instructions.
Consequently, OFA, with its capability to process bounding box coordinates and its moderate model size, serves as a suitable foundation for instruction tuning.

\dataset{} is divided into 8:1:1 for training, validation, and test set.
The AdamW optimizer~\citep{adamw} with \( \beta_1=0.9 \) and \( \beta_2=0.999 \) is utilized for optimization. 
To avoid overfitting, we apply regularization techniques \ie, dropout with a rate of \( 0.1 \), weight decay of \( 0.01 \), and label smoothing set at \( 0.1 \). 
We implement a linear decay learning rate scheduler with an initial warmup ratio of \( 0.06 \), and \model{} is trained on 4 NVIDIA RTX A6000 GPUs for a total of \(3\) epochs.




\begin{table*}[ht]
\caption{Detailed hyperparameters of \model{} configuration. We list the configuration for \model{} of 4 different sizes.
}
\center
\begin{adjustbox}{max width=1.\textwidth}
\begin{tabular}{lccccccc}
\toprule
  Model
  &\#Param.
  &Backbone
  &Hidden size
  &Intermediate Size
  &\#Head
  &\#Enc. Layers
  &\#Dec. Layers
  \\
\midrule
  \model{}$_{\text{tiny}}$
  &33M
  &ResNet50
  &256
  &1024
  &4
  &4
  &4
  \\
  \model{}$_{\text{medium}}$
  &93M
  &ResNet101
  &512
  &2048
  &8
  &4
  &4
  \\
  \model{}$_{\text{base}}$
  &182M
  &ResNet101
  &768
  &3072
  &12
  &6
  &6
  \\
  \model{}$_{\text{large}}$
  &472M
  &ResNet152
  &1024
  &4096
  &16
  &12
  &12
  \\
\bottomrule
\end{tabular}
\end{adjustbox}
\label{table:model_configuration}
\end{table*}

\subsection{Output with Multiple Triples}

We model \model{} in a generative manner, allowing the generation of novel commonsense triples that are not contained in the training set.
Crucially, \dataset{} provides a rich set of commonsense triples for each object-type pair.
we devise a strategy to preserve this diversity in the generated output.

For seen commonsense, given \(n\) triples of an object \(o\) and a type \(r\), \(\{(o, r, ec_{1}), ..., (o, r, ec_{n})\}\), we sample \(m\) triples and join their tail nodes with \([sep]\), yielding \(ec_{1}[sep] ... [sep]ec_{m}\), where \(m \leq n\) helps manage the number of commonsense triples desired during generation.

Given the sorted unseen commonsense as described in Sec.~\ref{ssec:mapping_region_phrases}, we concatenate the top-\(k\) tail nodes into \(ic_{1}[sep] ... ic_{k}\) and then add \(j\) random nodes sampled from the remaining to form \(ic_{1}[sep] ... ic_{k}[sep]ics_{k+1}, ... ics_{k+j}\).
This strategy ensures the generation of high-priority unseen commonsense while also maintaining a diversity of lower-priority commonsense.

\subsection{Evaluation Details of \model{}}
We select powerful open and closed source MLLMs on MME leaderboard~\cite{mme}, SPHINX, Qwen-VL-7B, and GPT-4o~\cite{gpt4}, and then manually compare them with \model{}$_{\text{large}}$ by carefully crafting the prompts.

The crafted prompts and qualitative results are illustrated in Tab.~\ref{table:prompt_gpt4v} and Tab.~\ref{table:prompt_SPHINX}.
Texts in double quotations are key components that can be adapted based on the object under consideration, the type of visual commonsense, \etc.
Here, we show only the optimally chosen prompt that was used for the experimental results in Tab.~\ref{table:human_evaluation}.
\begin{table*}[t]
\caption{Detailed prompts for evaluating GPT-4o.}
\center
\begin{adjustbox}{max width=1.\textwidth}
\begin{tabular}{@{}ll@{}}
\toprule
\multicolumn{1}{c}{\textbf{Type}} &
  \multicolumn{1}{c}{\textbf{Prompt}} \\ \midrule
\textit{/E/P/HasProperty} &
  \begin{tabular}[c]{@{}l@{}}{[}image{]} List some ``visible'' ``has property'' commonsense about the ``field'' located in the red bounding box in the image.\\ Output in the following format: (filed, visible has property, short phrase).\end{tabular} \\
\textit{/E/S/LocatedNear} &
  \begin{tabular}[c]{@{}l@{}}{[}image{]} List some ``visible'' ``located near'' commonsense about the ``cheese'' located in the red bounding box in the image.\\ Output in the following format: (cheese, visible located near, short phrase).\end{tabular} \\
\textit{/E/S/Relatedness} &
  \begin{tabular}[c]{@{}l@{}}{[}image{]} List some ``visible'' ``related spatial relation'' commonsense about the ``hamburger'' located in the red bounding box in\\ the image. Output in the following format: (hamburger, related spatial relation, on the desk).\end{tabular} \\
\textit{/E/A/CapableOf} &
  \begin{tabular}[c]{@{}l@{}}{[}image{]} List some ``visible'' ``capable of'' commonsense about the ``wings'' located in the red bounding box in the image.\\ Output in the following format: (wings, visible capable of, short phrase).\end{tabular} \\
\textit{/E/A/ReceivesAction} &
  \begin{tabular}[c]{@{}l@{}}{[}image{]} List some ``visible'' ``receives passive action'' commonsense about the ``window'' located in the red bounding box\\ in the image. Output in the following format: (window, receives passive action, short phrase).\end{tabular} \\
\textit{/I/P/HasProperty} &
  \begin{tabular}[c]{@{}l@{}}{[}image{]} List some ``invisible'' ``has property'' commonsense about the ``hose'' located in the red bounding box in the image.\\ Output in the following format: (hose, invisible has property, short phrase).\end{tabular} \\
\textit{/I/P/CreatedBy} &
  \begin{tabular}[c]{@{}l@{}}{[}image{]} List some ``invisible'' ``created by'' commonsense about the ``tree'' located in the red bounding box in the image.\\ Output in the following format: (hose, invisible created by, short phrase).\end{tabular} \\
\textit{/I/S/LocatedNear} &
  \begin{tabular}[c]{@{}l@{}}{[}image{]} List some ``invisible'' ``located near'' commonsense about the ``shadow'' located in the red bounding box in the image.\\ Output in the following format: (shadow, invisible located near, short phrase).\end{tabular} \\
\textit{/I/A/CapableOf} &
  \begin{tabular}[c]{@{}l@{}}{[}image{]} List some ``invisible'' ``capable of'' commonsense about the ``tire'' located in the red bounding box in the image.\\ Output in the following format: (tire, invisible capable of, short phrase).\end{tabular} \\
\textit{/I/A/UsedFor} &
  \begin{tabular}[c]{@{}l@{}}{[}image{]} List some ``invisible'' ``used for'' commonsense about the ``car'' located in the red bounding box in the image.\\ Output in the following format: (car, invisible used for, short phrase).\end{tabular} \\
\textit{/I/A/ReceivesAction} &
  \begin{tabular}[c]{@{}l@{}}{[}image{]} List some ``invisible'' ``receives passive action'' commonsense about the ``picture'' located in the red bounding box\\ in the image. Output in the following format: (picture, invisible receives passive action, short phrase).\end{tabular} \\ \bottomrule
\end{tabular}
\end{adjustbox}
\label{table:prompt_gpt4v}
\end{table*}
\begin{table*}[t]
\caption{Detailed prompts for evaluating SPHINX and Qwen-VL-7B.}
\center
\begin{adjustbox}{max width=1.\textwidth}
\begin{tabular}{@{}ll@{}}
\toprule
\multicolumn{1}{c}{\textbf{Type}} &
  \multicolumn{1}{c}{\textbf{Prompt}} \\ \midrule
\textit{/E/P/HasProperty} &
  \begin{tabular}[c]{@{}l@{}}{[}image{]} Can you tell me what visible properties common sense the ``field'' has in this image?\end{tabular} \\
\textit{/E/S/LocatedNear} &
  \begin{tabular}[c]{@{}l@{}}{[}image{]} Can you tell me what exists near this ``cheese'' in the bounding box of the image?\end{tabular} \\
\textit{/E/S/Relatedness} &
  \begin{tabular}[c]{@{}l@{}}{[}image{]} Can you tell me anything about the location of the ``hamburger'' in the picture? \end{tabular} \\
\textit{/E/A/CapableOf} &
  \begin{tabular}[c]{@{}l@{}}{[}image{]} Please list some ``capable of'' commonsense you can see about ``wings'' in the bounding box of the image.\end{tabular} \\
\textit{/E/A/ReceivesAction} &
  \begin{tabular}[c]{@{}l@{}}{[}image{]} Can you tell me what are the common sense passively accepted actions of the ``window'' in the picture?\end{tabular} \\
\textit{/I/P/HasProperty} &
  \begin{tabular}[c]{@{}l@{}}{[}image{]} Can you tell me what visible properties commonsense the ``car'' has in this image?\end{tabular} \\
\textit{/I/P/CreatedBy} &
  \begin{tabular}[c]{@{}l@{}}{[}image{]} Please list some ``created by'' commonsense you can imagine about ``tree'' in the bounding box in the image.\end{tabular} \\
\textit{/I/S/LocatedNear} &
  \begin{tabular}[c]{@{}l@{}}{[}image{]} Please list some ``located near'' commonsense you can imagine about ``shadow'' in the bounding box of the image.\end{tabular} \\
\textit{/I/A/CapableOf} &
  \begin{tabular}[c]{@{}l@{}}{[}image{]} Please list some ``capable of'' commonsense you can imagine about ``tire'' in the bounding box of the image.\end{tabular} \\
\textit{/I/A/UsedFor} &
  \begin{tabular}[c]{@{}l@{}}{[}image{]} Please use your imagination to tell me what can ``car'' in the picture be used for.\end{tabular} \\
\textit{/I/A/ReceivesAction} &
  \begin{tabular}[c]{@{}l@{}}{[}image{]} Please list some ``receives passive action'' commonsense you can imagine about ``picture'' in the bounding box\\ of the image.\end{tabular} \\ \bottomrule
\end{tabular}
\end{adjustbox}
\label{table:prompt_SPHINX}
\end{table*}

\subsection{Qualitative Results of \model{}}
\label{sec:qualitative}

\begin{figure*}[t]
  \centering
   \includegraphics[width=1.\linewidth]{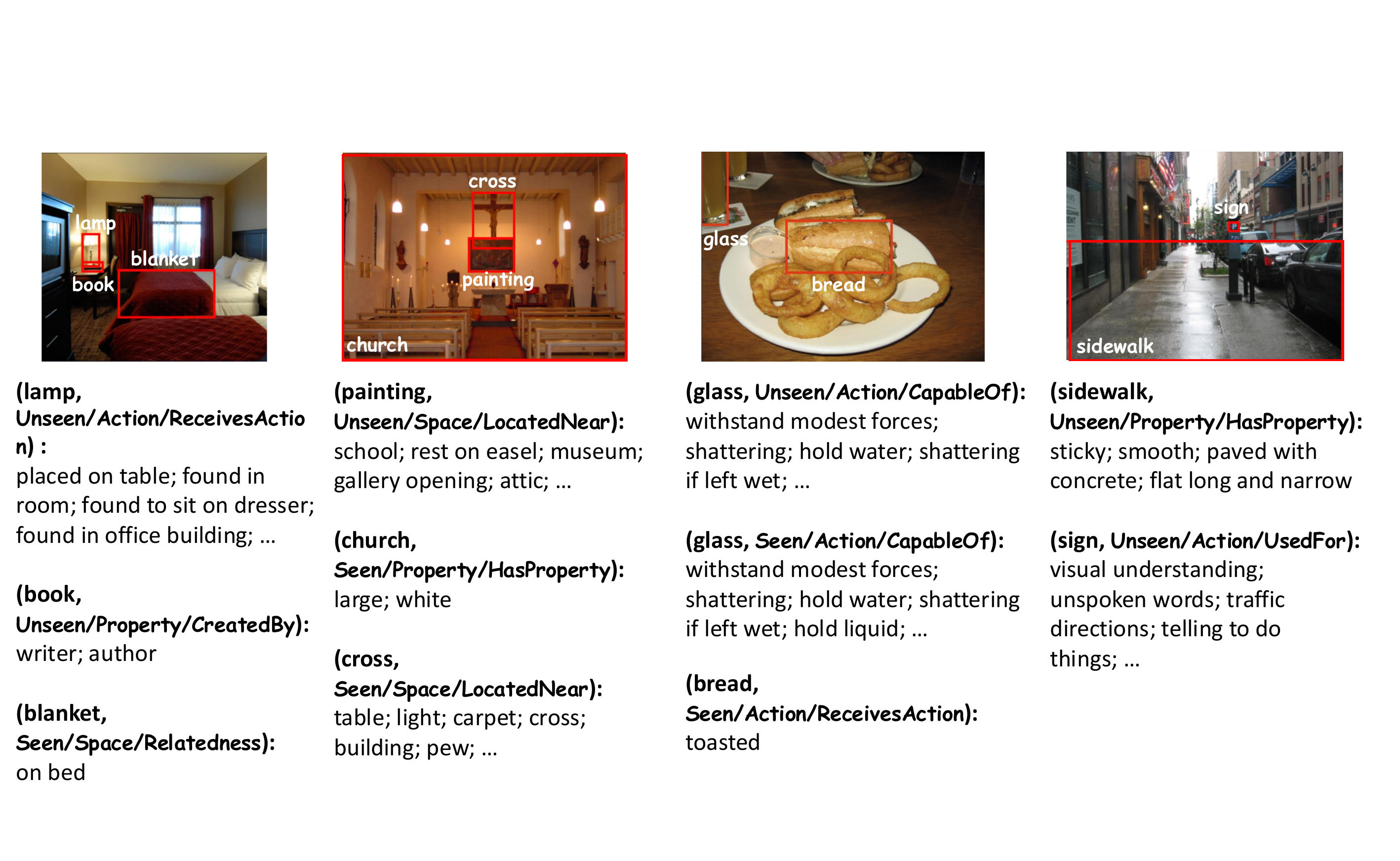}
   \caption{Qualitative results generated by \model{}$_{\text{large}}$.}
   \label{fig:examples}
\end{figure*}

In this section, we showcase a variety of examples produced by \model{}$_{\text{large}}$.
As observed in Fig.~\ref{fig:examples}, it is evident that \model{}$_{\text{large}}$ is capable of generating high-quality representations of both seen and unseen commonsense categories associated with a specific object in an image.
While the results are generally accurate, there may be occasional errors.
For instance, in the final image of Fig.~\ref{fig:examples}, ``\textit{/Unseen/Property/HasProperty}'' of ``sidewalk'' is correctly discovered as ``paved with concrete'';
however, this description could be more precisely attributed to the type of ``\textit{/Seen/Action/ReceivesAction}''.

\section{Details of Downstream VL Tasks}
\subsection{Details of Evaluation on \imagenetvc{}}

\imagenetvc{} evaluates commonsense understanding across multiple dimensions, including color, shape, material, components, and other attributes of various objects. For example, the question “What is the color of a koala?” from \imagenetvc{} assesses the model’s knowledge of a koala’s typical color, which is brown. To evaluate performance on \imagenetvc{}, we compare the backbone model with the backbone trained on \dataset{} by directly testing them on \imagenetvc{}, using experimental setting provided by \citep{imagenetvc}.

\subsection{Details of Evaluation on VQA}

We evaluates on VQA tasks to underscore the significance of the commonsense discovered by \task{}.
While visual commonsense reasoning could be a good option, the datasets~\citep{visual_commonsense_reasoning, visual_comet} mostly focus on human behaviors and states, not offering a broad reflection of commonsense's role for diverse objects as discovered by \task{}.
Therefore, we select VQAv2~\citep{making} OK-VQA~\citep{okvqa} dataset,a visual question answering task requiring models to utilize visual information from images to answer questions.

It is reasonable to assume that performing \task{} on an image can provide additional insights that are instrumental in improving VQA performance, as commonsense serves as additional information for better question answering.

For each question in VQAv2 and OK-VQA, we begin by identifying the entities contained within the questions using dependency parsing.
Next, we employ OFA$_{\text{large}}$ to determine the bounding boxes corresponding to these entities in the image.
Finally, we used \model{}$_{\text{large}}$ for visual commonsense discovery pertaining to the identified objects in the image.

It is important to note that not all the commonsense discovered by \model{}$_{\text{large}}$ is equally beneficial for answering the questions.
As such, we filter the commonsense and retain only what is most relevant to the question.
The reserved commonsense is then concatenated with the question, enriching the context for the answer generation.

For fair comparison, both the backbone and the backbone with visual commonsense are finetuned on VQAv2 and OK-VQA from scratch under the identical experimental setting.

To validate the effectiveness of different types of commonsense, we randomly select 100 samples from VQA-v2 and evaluate the accuracy manually with different types of commonsense.
Tab.~\ref{table:vqa_human} shows the efficacy of seen over unseen commonsense for VQAv2, since VQAv2 is a relatively simple VQA dataset where most of the questions are related to seen commonsense.

\begin{table}[ht]
\center
\caption{Accuracy with different type of commonsense on VQAv2.}
\begin{adjustbox}{max width=1\columnwidth}
\begin{tabular}{@{}lccc@{}}
\toprule
 Category & \ding{56}   & Explicit & Implicit
 \\ \midrule
Accuracy & 0.89  & 0.92 & 0.90  \\ 
\bottomrule
\end{tabular}
\end{adjustbox}
\label{table:vqa_human}
\end{table}

\end{document}